\documentclass[journal]{IEEEtran}
\usepackage{stmaryrd}
\usepackage{bm}
\usepackage{booktabs}
\usepackage{graphicx}
\usepackage{cite}
\usepackage{float}
\usepackage{graphicx}
\usepackage{stfloats}
\usepackage{amsmath}
\usepackage{epstopdf}
\usepackage{multirow}
\usepackage[colorlinks,linkcolor=red,anchorcolor=green,citecolor=green]{hyperref}
\usepackage[numbers,sort&compress]{natbib}

\ifCLASSINFOpdf

\else

\fi

\begin{document}

\title{Non-Convex Weighted $\ell_p$ Minimization based Group Sparse Representation Framework for Image Denoising}

\author{Qiong~Wang, Xinggan~Zhang, Yu~Wu, Lan~Tang and Zhiyuan~Zha
\thanks{Q. Wang, X. Zhang, Y. Wu and Z. Zha are with the department of Electronic Science and Engineering,
Nanjing University, Nanjing 210023, China. E-mail: zhazhiyuan.mmd@gmail.com.}
\thanks{L. Tang is the department of Electronic Science and Engineering, Nanjing University,
and National Mobile Commun. Research Lab., Southeast University, Nanjing 210023, China.}
\thanks{This work was supported by the NSFC (61571220, 61462052, 61502226) and the open research fund of National Mobile Commune. Research Lab., Southeast University (No.2015D08).}}


\maketitle

\begin{abstract}
Nonlocal image representation or group sparsity has attracted considerable interest in various low-level vision tasks and has led to several state-of-the-art image denoising techniques, such as BM3D, LSSC. In the past, convex optimization with sparsity-promoting convex regularization was usually regarded as a standard scheme for estimating sparse signals in noise. However, using convex regularization cannot still obtain the correct sparsity solution under some practical problems including image inverse problems. In this paper we propose a non-convex weighted $\ell_p$ minimization based group sparse representation (GSR) framework for image denoising. To make the proposed scheme tractable and robust, the generalized soft-thresholding (GST) algorithm is adopted to solve the non-convex $\ell_p$ minimization problem. In addition, to improve the accuracy of the nonlocal similar patch selection, an adaptive patch search (APS) scheme is proposed. Experimental results demonstrate that the proposed approach not only outperforms many state-of-the-art denoising methods such as BM3D and WNNM, but also results in a competitive speed.
\end{abstract}


\IEEEpeerreviewmaketitle

\section{Introduction}

\IEEEPARstart{T}{he} goal of image denoising is to restore the  clean image $\textbf{\emph{X}}$ from its noisy observation $\textbf{\emph{Y}}$ as accurately as possible, while preserving significant detail features such as edges and textures. The degradation model for the denoising problem can be represented as: $\textbf{\emph{Y}}=\textbf{\emph{X}} +\textbf{\emph{V}}$, where $\textbf{\emph{V}}$ is usually assumed to be additive white Gaussian noise. Image denoising problem is mathematically ill-posed and image priors are exploited to adjust it such that meaningful solutions exist. Over the past few decades, numerous image denoising methods have been developed, including total variation based \cite{1,2}, sparse representation based \cite{3,4}, nonlocal self-similarity based \cite{5,6,7,8} and deep learning based ones \cite{10,11,40}, etc.

Early models mainly consider the priors on level of pixel, such as  total variation (TV) regularization methods \cite{1,2}.  These methods actually assume that natural image gradients exhibit heavy-tailed distributions, which can be fitted by Laplacian or hyper-Laplacian models \cite{12}. Since the TV model favors the piecewise constant image structures, it often damages the image details and tends to over-smooth the images.

As an alternative, another significant property of natural images is to model the prior on patches. The most representative work is sparse representation based scheme \cite{3,4}, which encodes an image patch as a sparse linear combination of the atoms in an over-complete redundant dictionary. The dictionary is usually learned from  natural images \cite{13}. The seminal of KSVD dictionary \cite{4} has not only confirmed promising denoising performance, but also extended and successfully exploited it in various image processing and computer vision tasks \cite{14,15}.  However, patch-based sparse representation model usually suffers from some limits, such as dictionary learning with great computational complexity and  neglecting the relationships among similar patches \cite{7,16,17}.

Motivated by the observation that nonlocal similar patches in a natural image are linearly correlated with each other, this so-called nonlocal self-similarity (NSS) prior was initially employed in the work of nonlocal means denoising \cite{5}, which has become the most effective priors for the task of image restoration \cite{18,19}. Due to its favorable reconstruction performance, a large amount of further developments have been proposed \cite{6,7,8,16,17,20,43}. For instance, a very popular scheme is BM3D \cite{6}, which groups similar patches into 3D array and disposes these arrays by sparse collaborative filtering. Marial $\emph{et al}.$  \cite{7} proposed the learned simultaneous sparse coding (LSSC) to improve the denoising performance of K-SVD \cite{4} via group sparse coding. Gu $\emph{et al}.$ \cite{20,21} proposed the weighted nuclear norm minimization (WNNM) model, which turned the image denoising into the problem of low rank matrix approximation of noisy nonlocal similar patches. Lately, deep learning based techniques for image denoising have been attracting considerable attentions due to its impressive denoising performance \cite{10,11,40}.

Traditional sparse representation based image denoising methods exploit the $\ell_1$-norm based sparsity of an image and the resulting convex optimization problems can be efficiently solved by the class of surrogate-function based methods \cite{22,23}. However, using convex regularization cannot still obtain the correct sparsity solution under some practical problems including image inverse problems \cite{41}.

Inspired by the success of $\ell_p$ ($0<p<1$) sparse optimization \cite{24,25,26,42} and our previous work \cite{41}, this paper proposes a non-convex weighted $\ell_p$ minimization based group sparse representation (GSR) framework for image denoising. To make the proposed scheme tractable and robust, the generalized soft-thresholding (GST) algorithm is adopted to solve the non-convex $\ell_p$ minimization problem. Moreover, we propose an adaptive patch search (APS) scheme to improve the accuracy of the nonlocal similar patch selection. Experimental results  show that the proposed approach not only outperforms many state-of-the-art denoising methods such as BM3D and WNNM, but also results in a competitive speed.
\section{Group-based Sparse Representation}
Recent advances have suggested that structured or group sparsity can offer powerful  performance for image restoration \cite{7,8,17}. Since the unit of our proposed sparse representation model is group, this section will give briefs to introduce how to construct the groups. More specifically,  image  $\textbf{\emph{X}}$ with size $\emph{N}$ is divided into $\emph{n}$ overlapped patches $\textbf{\emph{x}}_i$ of size $\sqrt{d}\times\sqrt{d}, i=1,2,...,n$.  Then for each exemplar patch $\textbf{\emph{x}}_i$, its most similar $m$ patches are selected from an $L \times L$ sized searching window to form a set ${\textbf{\emph{S}}}_i$. Since then, all the patches in ${\textbf{\emph{S}}}_i$ are stacked into a matrix ${\textbf{\emph{X}}}_i\in\Re^{{d}\times {m}}$, which contains every element of ${\textbf{\emph{S}}}_i$ as its column, i.e., ${\textbf{\emph{X}}}_i=\{{\textbf{\emph{x}}}_{i,1}, {\textbf{\emph{x}}}_{i,2}, ..., {\textbf{\emph{x}}}_{i,m}\}$.  The matrix ${\textbf{\emph{X}}}_i$ consisting of  all the patches with similar structures is called as a group, where ${\textbf{\emph{x}}_{i,m}}$ denotes the $m$-th similar patch (column form) of the $i$-th group. Finally, similar to patch-based sparse representation \cite{3,4}, given a dictionary ${\textbf{\emph{D}}}_i$, which is often learned from each group, such as DCT, PCA-based dictionary \cite{34}, each group ${\textbf{\emph{X}}}_i$ can be sparsely represented as $\boldsymbol\alpha_i={{\textbf{\emph{D}}}_i}^{-1}\textbf{\emph{X}}_i$ and solved by the following $\ell_0$-norm minimization problem,
\begin{equation}
\boldsymbol\alpha_i=\arg\min_{\boldsymbol\alpha_i}\sum\nolimits_{i=1}^n (\frac{1}{2}||{\textbf{\emph{X}}}_i-{\textbf{\emph{D}}}_i\boldsymbol\alpha_i||_F^2+\lambda_i||\boldsymbol\alpha_i||_0)
\label{eq:1}
\end{equation} 
where  $||\bullet||_F^2$ denotes the Frobenious norm and $\lambda_i$ is the regularization parameter. $||\bullet||_0$ is $\ell_0$-norm, counting the nonzero entries of $\boldsymbol\alpha_i$.

In image denoising,  each noise patch $\textbf{\emph{y}}_i$ is extracted from the  noisy image ${\textbf{\emph{Y}}}$. We search for its similar patches to generate a group ${\textbf{\emph{Y}}}_i$, i.e., ${\textbf{\emph{Y}}}_i=\{{\textbf{\emph{y}}}_{i,1}, {\textbf{\emph{y}}}_{i,2},...,{\textbf{\emph{y}}}_{i,m}\}$. Thus, image denoising  is translated into how to reconstruct ${\textbf{\emph{X}}}_i$ from ${\textbf{\emph{Y}}}_i$ by using group sparse representation,
\begin{equation}
\boldsymbol\alpha_i=\arg\min_{\boldsymbol\alpha_i}\sum\nolimits_{i=1}^n (\frac{1}{2}||{\textbf{\emph{Y}}}_i-{\textbf{\emph{D}}}_i\boldsymbol\alpha_i||_F^2+\lambda_i||\boldsymbol\alpha_i||_0)
\label{eq:2}
\end{equation} 

Once all group sparse codes $\{\boldsymbol\alpha_i\}$ are obtained, the latent clean image $\textbf{\emph{X}}$ can be reconstructed as $\hat{\textbf{\emph{X}}}={\textbf{\emph{D}}}\boldsymbol\alpha$, where the group sparse code $\boldsymbol\alpha$ includes the set of $\{\boldsymbol\alpha_i\}$.

However, since the $\ell_0$ minimization is discontinuous optimization and NP-hard, solving Eq.~\eqref{eq:2} is a difficult combinatorial optimization problem. For this reason, it has been suggested that  $\ell_0$ minimization can be replaced by its convex $\ell_1$ counterpart,
\begin{equation}
\boldsymbol\alpha_i=\arg\min_{\boldsymbol\alpha_i} \sum\nolimits_{i=1}^n (\frac{1}{2}||{\textbf{\emph{Y}}}_i-{\textbf{\emph{D}}}_i\boldsymbol\alpha_i||_F^2+\lambda_i||\boldsymbol\alpha_i||_1)
\label{eq:3}
\end{equation} 

However,  $\ell_1$ minimization is hard to achieve the desired sparsity solution in some practical problems, such as image denoising, image compressive sensing \cite{28,29}, etc.
\section{Non-convex Weighted $\ell_p$ minimization based Group Sparse Representation Framework for Image Denoising}
Conventional convex optimization with sparsity-promoting convex regularization is usually regarded as a standard scheme for estimating sparse signals in noise. However, using convex regularization cannot still obtain the correct sparsity solution under some practical problems including image inverse problems \cite{41}. This section introduces a non-convex weighted $\ell_p$ minimization based group sparse representation framework for image denoising. To make the optimization tractable, the generalized soft-thresholding (GST) algorithm \cite{26} is adopted to solve the non-convex $\ell_p$ minimization problem. To improve the accuracy of the nonlocal similar patch selection, an adaptive patch search scheme is proposed.
\subsection{Modeling of Non-convex Weighted $\ell_p$ Minimization}
Inspired by the success of $\ell_p$ ($0<p<1$) sparse optimization \cite{24,25,26,42} and our previous work \cite{41}, to obtain sparsity solution more accurately, we extend the non-convex weighted $\ell_p$ ($0<p<1$) penalty function on group sparse coefficients of the data matrix to substitute the convex $\ell_1$ norm. Specifically, instead of Eq.~\eqref{eq:3}, a non-convex weighted $\ell_p$ minimization based group sparse representation framework for image denoising is proposed by solving the following minimization,
\begin{equation}
\boldsymbol\alpha_i=\arg\min_{\boldsymbol\alpha_i}\sum\nolimits_{i=1}^n
(\frac{1}{2}||{\textbf{\emph{Y}}}_i-{\textbf{\emph{D}}}_i\boldsymbol\alpha_i||_F^2+||{{\textbf{\emph{W}}}_i}\boldsymbol\alpha_i||_p)
\label{eq:4}
\end{equation} 
where ${{\textbf{\emph{W}}}_i}$ is a weight assigned to each group ${\textbf{\emph{Y}}}_i$. Each weight matrix ${{\textbf{\emph{W}}}_i}$ will enhance the representation capability of each group sparse coefficient $\boldsymbol\alpha_i$. In addition, one important issue of the proposed denoising approach is the selection of the dictionary. To adapt to the local image structures, instead of learning an over-complete dictionary for each group ${{\textbf{\emph{Y}}}}_i$ as in \cite{7}, we learn the principle component analysis (PCA) based dictionary \cite{34} for each group ${{\textbf{\emph{Y}}}}_i$. Due to orthogonality of each dictionary $\textbf{\emph{D}}_i$, and thus, based on the orthogonal invariance, Eq.~\eqref{eq:4} can be rewritten as
\begin{equation}
\begin{aligned}
&{{\boldsymbol\alpha}}_i=\min\limits_{{{{\boldsymbol\alpha}}}_i}\sum\nolimits_{i=1}^n(\frac{1}{2}||{{{{\boldsymbol\gamma}}}_i}-{{{{\boldsymbol\alpha}}}_i}||_F^2
 +||{\textbf{\emph{W}}}_i\boldsymbol\alpha_i||_p)\\
&=\min\limits_{\tilde{{{\boldsymbol\alpha}}}_i}\sum\nolimits_{i=1}^n(\frac{1}{2}||{\tilde{{{\boldsymbol\gamma}}}_i}-{\tilde{{{\boldsymbol\alpha}}}_i}||_2^2
 +||{\tilde{\textbf{\emph{w}}}}_i \tilde{\boldsymbol\alpha}_i||_p)\\
 \end{aligned}
\label{eq:5}
\end{equation}
where ${\textbf{\emph{Y}}}_i={{\textbf{\emph{D}}}_i{{{\boldsymbol\gamma}}}_i}$. ${\tilde{{{\boldsymbol\alpha}}}_i}$, ${\tilde{{{\boldsymbol\gamma}}}_i}$ and ${\tilde{\textbf{\emph{w}}}}_i$ denote the vectorization of the matrix ${{{{\boldsymbol\alpha}}}_i}$, ${{{{\boldsymbol\gamma}}}_i}$ and ${\textbf{\emph{W}}}_i$, respectively.
\subsection{Solving the Non-convex Weighted $\ell_p$ Minimization by the Generalized Soft-thresholding Algorithm }
To achieve the solution of Eq.~\eqref{eq:5} effectively, in this subsection, the generalized soft-thresholding (GST) algorithm \cite{26} is used to solve Eq.~\eqref{eq:5}. Specifically, given $p$, ${\tilde{{{\boldsymbol\gamma}}}_i}$ and ${\tilde{\textbf{\emph{w}}}}_i$, there exists a specific threshold,
\begin{equation}
\tau_p^{\emph{GST}}({\tilde{{\emph{w}}}_{i,j}})=(2{\tilde{{\emph{w}}}_{i,j}}(1-p))^{\frac{1}{2-p}}+{\tilde{{\emph{w}}}_{i,j}}p(2{\tilde{{\emph{w}}}_{i,j}}(1-p))^{\frac{p-1}{2-p}}
\label{eq:6}
\end{equation} 
where ${\tilde{\gamma}_{i,j}}$, ${\tilde{\alpha}_{i,j}}$ and ${\tilde{{\emph{w}}}_{i,j}}$ are the $j$-th element of  ${\tilde{{{\boldsymbol\gamma}}}_i}$, ${\tilde{{{\boldsymbol\alpha}}}_i}$ and  ${\tilde{\textbf{\emph{w}}}}_i$, respectively. Here if ${\tilde{\gamma}_{i,j}}<\tau_p^{\emph{GST}}({\tilde{{\emph{w}}}_{i,j}})$, ${\tilde{\alpha}_{i,j}}=0$ is the global minimum. Otherwise, the optimum will be obtained at non-zero point. According to \cite{26}, for any ${\tilde{\gamma}_{i,j}}\in(\tau_p^{\emph{GST}}({\tilde{{\emph{w}}}_{i,j}}), +\infty)$, Eq.~\eqref{eq:5} has one unique minimum ${\textbf{\emph{T}}}_p^{\emph{GST}}({\tilde{\gamma}_{i,j}}; {\tilde{{\emph{w}}}_{i,j}})$, which can be obtained by solving the following equation,
\begin{equation}
{\textbf{\emph{T}}}_p^{\emph{GST}}({\tilde{\gamma}_{i,j}}; {\tilde{{\emph{w}}}_{i,j}})- {\tilde{\gamma}_{i,j}} + {\tilde{{\emph{w}}}_{i,j}}p \left({\textbf{\emph{T}}}_p^{\emph{GST}}({\tilde{\gamma}_{i,j}}; {\tilde{{\emph{w}}}_{i,j}})\right)^{p-1} =0
\label{eq:7}
\end{equation} 

The complete description of the GST algorithm is exhibited in Algorithm 1.  For more details about the GST algorithm, please refer to \cite{26}.
\begin{table}[!htbp]
\centering  
\begin{tabular}{lccc}  
\hline  
\qquad \ \  Algorithm 1: Generalized Soft-Thresholding (GST) \cite{26}.\\
\hline
$\textbf{Input:}$ \ ${\tilde{\gamma}_{i,j}}, {\tilde{{\emph{w}}}_{i,j}}, p, J$.\\
1.\ \ \ $\tau_p^{\emph{GST}}({\tilde{{\emph{w}}}_{i,j}})=(2{\tilde{{\emph{w}}}_{i,j}}(1-p))^{\frac{1}{2-p}}+{\tilde{{\emph{w}}}_{i,j}}p(2{\tilde{{\emph{w}}}_{i,j}}(1-p))^{\frac{p-1}{2-p}}
$;\\
2.\ \ \ $\textbf{If}$ \ \ \ $|{\tilde{\gamma}_{i,j}}|\leq \tau_p^{\emph{GST}}({\tilde{{\emph{w}}}_{i,j}})$\\
3.\ \ \ \ \ \ \ \ \ ${\textbf{\emph{T}}}_p^{\emph{GST}}({\tilde{\gamma}_{i,j}}; {\tilde{{\emph{w}}}_{i,j}})=0$;\\
4.\ \ \ $\textbf{else}$\\
5.\ \ \ \ \ \ \ \ \ $k=0, {\tilde{\alpha}_{i,j}}^{(k)}=|{\tilde{\gamma}_{i,j}}|$;\\
6.\ \ \ \ \ \ \ \ \ Iterate on $k=0, 1, ..., J$\\
7.\ \ \ \ \ \ \ \ \ ${\tilde{\alpha}_{i,j}}^{(k+1)}=|{\tilde{\gamma}_{i,j}}|-{{\tilde{\emph{w}}}_{i,j}}p \left({\tilde{\alpha}_{i,j}}^{(k)}\right)^{p-1}$;\\
8.\ \ \ \ \ \ \ \ \ $k\leftarrow k+1$;\\
9.\ \ \ \ \ \ \ \ \ ${\textbf{\emph{T}}}_p^{\emph{GST}}({\tilde{\gamma}_{i,j}}; {\tilde{{\emph{w}}}_{i,j}})={\rm sgn}({\tilde{\gamma}_{i,j}}){\tilde{\alpha}_{i,j}}^{k}$;\\
10.\ \  $\textbf{End}$\\
$\textbf{Input:}$: ${\textbf{\emph{T}}}_p^{\emph{GST}}({\tilde{\gamma}_{i,j}}; {\tilde{{\emph{w}}}_{i,j}})$.\\
\hline
\end{tabular}
\end{table}
\subsection{Adaptive Patch Search}
$k$ Nearest Neighbors ($k$NN) method \cite{30} has been widely used to nonlocal similar patch selection. Given a noisy reference patch and a target dataset, the aim of $k$NN is to find the $k$ most similar patches. However, since the given reference patch is noisy, $k$NN has a drawback that some of the $k$ selected patches may not be truly similar to given reference patch. Therefore,  to obtain an effective similar patches index via $k$NN, an adaptive patch search scheme is proposed. We define the following formula,
\begin{equation}
\varphi ={\rm SSIM} (\boldsymbol\theta, {\hat{{\textbf{\emph{X}}}}}^{t+1})-{\rm SSIM} (\boldsymbol\theta, {\hat{{\textbf{\emph{X}}}}}^{t})
\label{eq:8}
\end{equation} 
where SSIM represents structural similarity \cite{31}, $\boldsymbol\theta$ is pre-filtering \footnote{This paper BM3D is chosen as a pre-filtering.} denoised image and ${\hat{{\textbf{\emph{X}}}}}^{t}$ represents the $t$-th iteration denoising result. We empirically define that if $\varphi<\rho$,  ${\hat{{\textbf{\emph{X}}}}}^{t+1}$ is regarded as target image to fetch the $k$ similar patch indexes of each group, otherwise  $\boldsymbol\theta$ is regarded as target image. $\rho$ is a small constant. 

For the weight $\textbf{\emph{W}}_i$ of each group sparse coefficient $\boldsymbol\alpha_i$, large values of each $\boldsymbol\alpha_i$ usually represent major edge and texture information. Therefore, we should shrink large values less, while shrinking smaller ones more \cite{32}. Inspired by \cite{33}, the weight $\textbf{\emph{W}}_i$ of each group $\textbf{\emph{Y}}_i$ is set as ${\tilde{\textbf{\emph{w}}}}_i= [{\tilde{{\emph{w}}}}_{i,1}, {\tilde{{\emph{w}}}}_{i,2}, ..., {\tilde{{\emph{w}}}}_{i,j}]$, where ${\tilde{{\emph{w}}}}_{i,j} =c*2\sqrt{2}\sigma^2/\boldsymbol\sigma_i$,  $\boldsymbol\sigma_i$ denotes the estimated variance of ${\tilde{\boldsymbol\alpha}_{i}}$, and $c$ is a small constant. 

In addition, we could execute the above denoising procedure for better results after several iterations. In the $t$-th iteration, the iterative regularization strategy \cite{35} is used to update the estimation of noise variance. Then the standard divation of noise in $t$-th iteration is adjusted as ${(\sigma^{t})}=\delta*\sqrt{({\sigma^2-||{{\textbf{\emph{Y}}}}-{\hat{{\textbf{\emph{X}}}}}^{t}||_2^2})}$, where $\delta$ is a constant. The proposed denoising procedure is summarized in Algorithm 2.
\begin{table}[!htbp]
\centering  
\begin{tabular}{lccc}  
\hline  
\qquad \ \  \ \ \ \ Algorithm 2: The Proposed Denoising Algorithm.\\
\hline
$\textbf{Input:}$ \ Noisy image ${{\textbf{\emph{Y}}}}$.\\
  $\rm \textbf{Initialization:} \  {\hat{{\textbf{\emph{X}}}}}={{\textbf{\emph{Y}}}}, \boldsymbol\theta, \emph{c}, \emph{d}, \emph{m}, \emph{L}, \emph{J}, \sigma, \rho, \delta, \lambda$;\\
  $\rm \textbf{For}$\ $t=1, 2, ..., K$ $\rm \textbf{do}$\\
  \qquad Iterative regularization ${{\textbf{\emph{Y}}}}^{t+1}= {\hat{\textbf{\emph{X}}}}^{t}+\lambda({{\textbf{\emph{Y}}}}-{\hat{\textbf{\emph{X}}}}^{t})$;\\
\qquad $\textbf{If}$ \ \ $t=1$\\
       \qquad \qquad Similar patch selection  based on $\boldsymbol\theta$.\\
\qquad $\textbf{Else}$\\
  \qquad\qquad $\textbf{If}$ \ \ ${\rm SSIM}({{\textbf{\emph{Y}}}}^{t+1}, \boldsymbol\theta)-{\rm SSIM}({{\textbf{\emph{Y}}}}^{t}, \boldsymbol\theta)<\rho$\\
  \qquad \qquad  \qquad Similar patches index selection  based on ${{\textbf{\emph{Y}}}}^{t+1}$.\\
  \qquad \qquad  $\textbf{Else}$\\
   \qquad \qquad \qquad Similar patches index selection  based on $\boldsymbol\theta$.\\
   \qquad  \qquad $\textbf{End if}$\\
\qquad $\textbf{End if}$\\
 \qquad $\rm \textbf{For}$\ each patch ${{\textbf{\emph{y}}}}_i$  $\rm \textbf{do}$\\
   \qquad \qquad  Find a group ${{{\textbf{\emph{Y}}}}_i}^{t+1}$ via $k$NN.\\
   \qquad \qquad Constructing dictionary ${{{\textbf{\emph{D}}}}_i}^{t+1}$ by ${{{\textbf{\emph{Y}}}}_i}$ by PCA operator.\\
    \qquad \qquad Generating the group sparse coefficient  ${\boldsymbol\gamma_i}^{t+1}$ by ${{{\textbf{\emph{D}}}}_i}^{-1}{{{\textbf{\emph{Y}}}}_i}$. \\
   \qquad \qquad Update ${\textbf{\emph{W}}_i}^{t+1}$ computing by ${\tilde{{\emph{w}}}}_{i,j} =c*2\sqrt{2}\sigma^2/\boldsymbol\sigma_i$.\\
   \qquad \qquad Update ${\boldsymbol\alpha_i}^{t+1}$ computing by Algorithm 1.\\
   \qquad \qquad Get the estimation ${{{\textbf{\emph{X}}}}_i}^{t+1}$ =${{{\textbf{\emph{D}}}}_i}^{t+1}$${\boldsymbol\alpha_i}^{t+1}$.\\
  \qquad  $\rm \textbf{End for}$\\
   \qquad \qquad Aggregate ${{{\textbf{\emph{X}}}}_i}^{t+1}$ to form the recovered image ${\hat{{\textbf{\emph{X}}}}}^{t+1}$.\\
    $\rm \textbf{End for}$\\
     $\textbf{Output:}$ ${\hat{\textbf{\emph{X}}}}^{t+1}$.\\
\hline
\end{tabular}
\label{lab:1}
\end{table}
\begin{figure}[!htbp]
\begin{minipage}[b]{1\linewidth}
  \centering
  \centerline{\includegraphics[width=9cm]{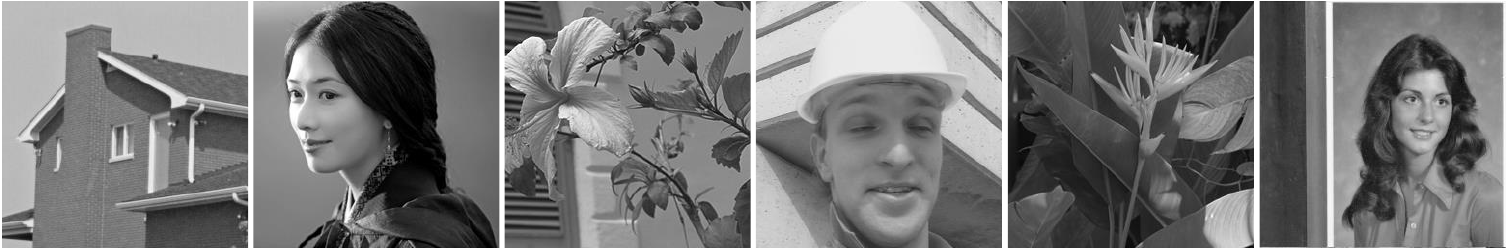}} 
\end{minipage}
\caption{The six test images for denoising experiments.}
\label{fig:1}
\end{figure}
\begin{figure}[!htbp]
\begin{minipage}[b]{1\linewidth}
  \centering
  \centerline{\includegraphics[width=9cm]{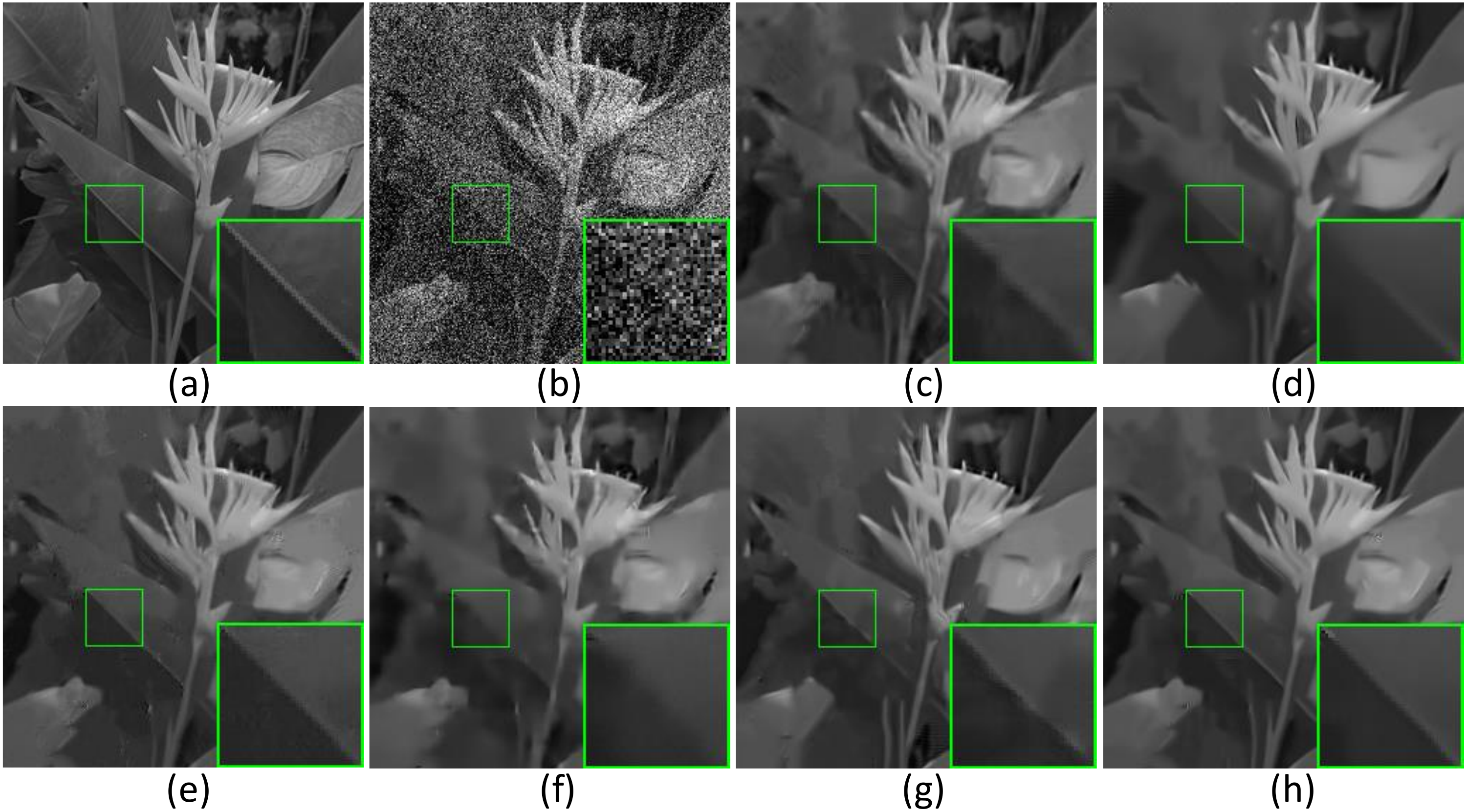}}
\end{minipage}
\caption{Denoising images of ${plants}$ by different methods ($\sigma=50$). (a) Original image; (b) Noisy image; (c) BM3D \cite{6} (PSNR=28.11dB); (d) LINC \cite{36} (PSNR=27.96dB); (e) AST-NLS \cite{37} (PSNR=28.04dB); (f) MSEPLL \cite{38} (PSNR=28.09dB); (g) WNNM \cite{21} (PSNR=28.23dB); (h) Proposed (PSNR=\textbf{28.60dB}).}
\label{fig:2}
\end{figure}

\section{Experimental Results}
To demonstrate the efficacy of the proposed denoising algorithm, in this section, we compare it with recently proposed state-of-the-art denoising methods, including BM3D \cite{6}, LINC \cite{36}, AST-NLS \cite{37}, MSEPLL \cite{38} and WNNM \cite{21}.  The experimental images are shown in Fig.~\ref{fig:1}. 
\textbf{The Matlab code can be downloaded at: \url{https://drive.google.com/open?id=0B0wKhHwcknCjM0doVFhlRElXWjg}}.
\begin{table*}[!htbp]
\caption{Denoising PSNR ($\textnormal{d}$B) results by different denoising methods.}
\scriptsize
\centering  
\begin{tabular}{|c||c|c|c|c|c|c||c|c|c|c|c|c|c|}
\hline
  \multicolumn{1}{|c||}{}&\multicolumn{6}{|c||}{$\sigma=20$}&\multicolumn{6}{|c|}{$\sigma=30$}\\
\hline
\multirow{1}{*}{\textbf{{Images}}}&{\textbf{{BM3D}}}&{\textbf{{LINC}}}&{\textbf{{AST-NLS}}}
&{\textbf{{MSEPLL}}}&{\textbf{{WNNM}}}&{\textbf{{Proposed}}}&{\textbf{{BM3D}}}
&{\textbf{{LINC}}}&{\textbf{{AST-NLS}}}&{\textbf{{MSEPLL}}}&{\textbf{{WNNM}}}&{\textbf{{Proposed}}}\\
 \hline
 \multirow{1}{*}{House}     & 33.77 &  33.82  & 33.87  & 33.27  & 34.04  & \textbf{34.08} & 32.09 & 32.26 & 32.26 & 31.71 & 32.52 & \textbf{32.65}\\
 \hline
 \multirow{1}{*}{lin}       & 32.83 &  33.04  & 33.84  & 32.80  & 33.00  & \textbf{33.08} & 30.95 & 31.03 & 30.83 & 30.96 & 31.07 & \textbf{31.14}\\
 \hline
 \multirow{1}{*}{flower}    & 30.01 &  30.30  & 30.28  & 30.10  & 33.34  & \textbf{30.48} & 27.97 & 28.13 & 28.20 & 28.05 & 28.26 & \textbf{28.36}\\
 \hline
 \multirow{1}{*}{foreman}   & 34.54 &  34.76  & 34.55  & 34.09  & 34.72  & \textbf{34.86} & 32.75 & 32.93 & 32.79 & 32.34 & 33.00 & \textbf{33.31}\\
 \hline
 \multirow{1}{*}{plants}    & 32.68 &  32.83  & 32.75  & 32.58  & 33.04  & \textbf{33.09} & 30.70 & 30.67 & 30.65 & 30.66 & 30.94 & \textbf{31.05}\\
 \hline
 \multirow{1}{*}{Miss}      & 33.71 &  33.64  & 33.64  & 33.68  & 33.70  & \textbf{33.80} & 31.89 & 31.75 & 31.72 & 31.92 & 31.93 & \textbf{32.04}\\
 \hline
 \multirow{1}{*}{\textbf{Average}}   & 32.92 &  33.07  & 32.99  & 32.80  & 33.14  & \textbf{33.23} & 31.06 & 31.13 & 31.08 & 30.93 & 31.29 & \textbf{31.42}\\
 \hline\hline
   \multicolumn{1}{|c||}{}&\multicolumn{6}{|c||}{$\sigma=40$}&\multicolumn{6}{|c|}{$\sigma=50$}\\
\hline
\multirow{1}{*}{\textbf{{Images}}}&{\textbf{{BM3D}}}&{\textbf{{LINC}}}&{\textbf{{AST-NLS}}}
&{\textbf{{MSEPLL}}}&{\textbf{{WNNM}}}&{\textbf{{Proposed}}}&{\textbf{{BM3D}}}
&{\textbf{{LINC}}}&{\textbf{{AST-NLS}}}&{\textbf{{MSEPLL}}}&{\textbf{{WNNM}}}&{\textbf{{Proposed}}}\\
 \hline
\multirow{1}{*}{House}     & 30.65 &  31.00  & 30.91  & 30.47  & 31.31  & \textbf{31.49} & 29.69 & 29.87 & 30.13 & 29.47 & 30.32 & \textbf{30.52}\\
 \hline
 \multirow{1}{*}{lin}      & 29.52 &  \textbf{29.94}  & 29.39  & 29.68  & 29.80  & 29.89 & 28.71 & 28.85 & 28.50 & 28.69 & 28.83 & \textbf{28.90}\\
 \hline
 \multirow{1}{*}{flower}   & 26.48 &  26.79  & 26.75  & 26.64  & 26.85  & \textbf{26.90} & 25.49 & 25.47 & 25.77 & 25.56 & 25.80 & \textbf{25.88}\\
 \hline
 \multirow{1}{*}{foreman}  & 31.29 &  31.31  & 31.29  & 31.05  & 31.54  & \textbf{32.08} & 30.36 & 30.33 & 30.46 & 30.04 & 30.75 & \textbf{31.03}\\
 \hline
 \multirow{1}{*}{plants}   & 29.14 &  29.09  & 29.05  & 29.25  & 29.28  & \textbf{29.70} & 28.11 & 27.96 & 28.04 & 28.09 & 28.23 & \textbf{28.60}\\
 \hline
 \multirow{1}{*}{Miss}     & 30.50 &  30.29  & 30.19  & 30.56  & 30.53  & \textbf{30.78} & 29.48 & 29.22 & 29.26 & 29.55 & 29.34 & \textbf{29.70}\\
 \hline
 \multirow{1}{*}{\textbf{Average}}     & 29.59 &  29.74  & 29.60  & 29.61  & 29.88  & \textbf{30.14} & 28.62 & 28.59 & 28.69 & 28.57 & 28.88 & \textbf{29.10}\\
 \hline
\end{tabular}
\label{lab:1}
\end{table*}

The parameter setting of proposed approach is as follows: the searching window $L \times L$ for similar patches is set to be  $30\times30$. The searching matched patches $m$ is set to be 60. The size of each patch $\sqrt{d} \times \sqrt{d}$ is set to be $6\times 6$ and $7\times 7$ for $\sigma\leq20$ and  $20<\sigma\leq50$, respectively. $(p, c, \lambda, \delta, \rho, J)$ are set to (1, 0.3, 0.1, 0.5,  2e-4, 2), (0.85, 0.3, 0.2, 0.8, 2e-4, 2), (0.8, 1.2, 0.1, 0.4, 6e-4, 2) and (0.75, 1.6, 0.1, 0.4, 2e-4, 2) for $\sigma\leq20, 20<\sigma\leq30, 30<\sigma\leq 40$ and $40<\sigma\leq 50$, respectively.
\begin{table}[!htbp]
\caption{Average PSNR ($\textnormal{d}$B) results of ADS and No-ADS  on 6 test images.}
\centering  
\begin{tabular}{|c|c|c|c|c|}
\hline
\multirow{1}{*}{$\sigma$}&{20}&{30}&{40}&{50}\\
 \hline
 \multirow{1}{*}{No-APS}      & 33.10 &  31.23 & 29.94 & 28.80 \\
 \hline
  \multirow{1}{*}{APS}  & \textbf{33.23} &  \textbf{31.42} & \textbf{30.14} & \textbf{29.10} \\
  \hline
\end{tabular}
\label{lab:2}
\end{table}

We first evaluate the proposed approach and the competing algorithms on 6 test images. Table~\ref{lab:1} shows the PSNR results. It can be seen that the proposed approach performs competitively compared to other methods. The proposed approach achieves 0.42dB, 0.34dB, 0.39dB, 0.51dB and 0.18dB improvement on average over the BM3D, LINC, AST-NLS, MSEPLL and WNNM, respectively. Fig.~\ref{fig:2} shows the denoised image of $\emph{plants}$ by the competing methods. It can be seen  that BM3D, LINC, AST-NLS, MSEPLL and WNNM still generate some undesirable artifacts and some details are lost. In contrast, the proposed approach not only  preserves the sharp edges, but also suppresses undesirable artifacts more effectively than  other competing methods.
\begin{table}[!htbp]
\caption{Average run time ($\textnormal{s}$) with different  methods on the 6 test images (size: $256\times 256$).}
\centering  
\begin{tabular}{|c|c|c|c|c|c|}
\hline
\multirow{1}{*}{Methods}&{LINC}&{AST-NLS}&{MSEPLL}&{WNNM}&{Ours}\\
 \hline
  \multirow{1}{*}{Average Time (s)}  & 263 &  300 & 182 & 172 & \textbf{82}\\
  \hline
\end{tabular}
\label{lab:3}
\end{table}

Second, to verify the proposed adaptive patch selection (APS) scheme effectively, we compare it with No-APS scheme. The average PSNR results of APS and No-APS schemes on 6 test images are shown in Table~\ref{lab:2}. One can observe that the PSNR results of APS scheme are better than No-APS. Thus, under the task of image denoising, the proposed APS scheme can enhance the accuracy of nonlocal similar patch selection.
\begin{table}[!htbp]
\caption{Average PSNR ($\textnormal{dB}$) results with different  methods on BSD200 dataset \cite{39}.}
\centering  
\begin{tabular}{|c|c|c|c|c|c|c|c|}
\hline
\multirow{1}{*}{$\sigma$}&{BM3D}&{LINC}&{AST-NLS}&{MSEPLL}&{WNNM}&{Ours}\\
 \hline
  \multirow{1}{*}{20}   & 29.86 & 29.92 &  29.98 & 29.95 & 30.11 & \textbf{30.14}\\
  \hline
  \multirow{1}{*}{30}   & 27.93 & 27.94 &  28.02 & 28.02 & \textbf{28.17} & 28.15\\
  \hline
  \multirow{1}{*}{40}   & 26.58 & 26.61 &  26.68 & 26.73 & 26.88 & \textbf{26.89}\\
  \hline
  \multirow{1}{*}{50}   & 25.71 & 25.64 &  25.80 & 25.84 & 25.96 & \textbf{25.97}\\
  \hline
\end{tabular}
\label{lab:4}
\end{table}

Third, to evaluate the computational cost of the competing algorithm, we compare the running time on 6 test images with different noise levels. All experiments are conducted under the Matlab 2012b environment on a machine with Intel (R) Core (TM) i3-4150 with 3.56Hz CPU and 4GB memory. The average run time (s) of the competing methods is shown in Table~\ref{lab:3}. It can be seen that the proposed approach clearly  requires less computation time than other methods. Note that the run time of the proposed approach includes the pre-filtering process.

Finally, We also comprehensively evaluate the proposed method on  200 test images from the BSD dataset \cite{39}. Table~\ref{lab:4} shows qualitative comparisons of the competing denosing methods on four noise levels ($\sigma=20, 30, 40, 50$). It can be seen that the proposed approach achieves very competitive denoising performance compared to WNNM.

\section{Conclusion}
Different from the conventional convex optimization, this paper proposed a non-convex weighted $\ell_p$ minimization based group sparse representation (GSR) framework for image denoising. To make the proposed scheme tractable and robust, we adopted the generalized soft-thresholding (GST) algorithm to solve the non-convex $\ell_p$ minimization problem. Moreover,  we proposed an adaptive patch search (APS) scheme to boost the accuracy of the nonlocal similar patch selection. Experimental results have verified  that the proposed approach outperforms many state-of-the-art denoising methods such as BM3D and WNNM, and results in a competitive speed.


{\footnotesize
\begin{thebibliography}{99}
\bibitem{1}
Rudin L I, Osher S, Fatemi E. Nonlinear total variation based noise removal algorithms[J]. Physica D: Nonlinear Phenomena, 1992, 60(1-4): 259-268.
\bibitem{2}
Chambolle A. An algorithm for total variation minimization and applications[J]. Journal of Mathematical imaging and vision, 2004, 20(1): 89-97.
\bibitem{3}
Elad M, Aharon M. Image denoising via sparse and redundant representations over learned dictionaries[J]. IEEE Transactions on Image processing, 2006, 15(12): 3736-3745.
\bibitem{4}
Aharon M, Elad M, Bruckstein A. $k$-SVD: An algorithm for designing overcomplete dictionaries for sparse representation[J]. IEEE Transactions on signal processing, 2006, 54(11): 4311-4322.
\bibitem{5}
Buades A, Coll B, Morel J M. A non-local algorithm for image denoising[C]//Computer Vision and Pattern Recognition, 2005. CVPR 2005. IEEE Computer Society Conference on. IEEE, 2005, 2: 60-65.
\bibitem{6}
Dabov K, Foi A, Katkovnik V, et al. Image denoising by sparse 3-D transform-domain collaborative filtering[J]. IEEE Transactions on image processing, 2007, 16(8): 2080-2095.
\bibitem{7}
Mairal J, Bach F, Ponce J, et al. Non-local sparse models for image restoration[C]//Computer Vision, 2009 IEEE 12th International Conference on. IEEE, 2009: 2272-2279.
\bibitem{8}
Zuo C, Jovanov L, Goossens B, et al. Image Denoising Using Quadtree-Based Nonlocal Means With Locally Adaptive Principal Component Analysis[J]. IEEE Signal Processing Letters, 2016, 23(4): 434-438.
\bibitem{10}
Liu S, Pan J, Yang M H. Learning recursive filters for low-level vision via a hybrid neural network[C]//European Conference on Computer Vision. Springer International Publishing, 2016: 560-576.
\bibitem{11}
Zhang K, Zuo W, Chen Y, et al. Beyond a Gaussian denoiser: Residual learning of deep CNN for image denoising[J]. IEEE Transactions on Image Processing, 2017.
\bibitem{12}
Zuo W, Zhang L, Song C, et al. Gradient histogram estimation and preservation for texture enhanced image denoising[J]. IEEE Transactions on Image Processing, 2014, 23(6): 2459-2472.
\bibitem{13}
Rubinstein R, Bruckstein A M, Elad M. Dictionaries for sparse representation modeling[J]. Proceedings of the IEEE, 2010, 98(6): 1045-1057.
\bibitem{14}
Zhang Q, Li B. Discriminative K-SVD for dictionary learning in face recognition[C]//Computer Vision and Pattern Recognition (CVPR), 2010 IEEE Conference on. IEEE, 2010: 2691-2698.
\bibitem{15}
Jiang Z, Lin Z, Davis L S. Label consistent K-SVD: Learning a discriminative dictionary for recognition[J]. IEEE Transactions on Pattern Analysis and Machine Intelligence, 2013, 35(11): 2651-2664.
\bibitem{16}
Dong W, Shi G, Li X. Nonlocal image restoration with bilateral variance estimation: A low-rank approach[J]. IEEE transactions on image processing, 2013, 22(2): 700-711.
\bibitem{17}
Zhang J, Zhao D, Gao W. Group-based sparse representation for image restoration[J]. IEEE Transactions on Image Processing, 2014, 23(8): 3336-3351.
\bibitem{18}
Katkovnik V, Foi A, Egiazarian K, et al. From local kernel to nonlocal multiple-model image denoising[J]. International journal of computer vision, 2010, 86(1): 1.
\bibitem{19}
Foi A, Boracchi G. Foveated nonlocal self-similarity[J]. International Journal of Computer Vision, 2016, 120(1): 78-110.
\bibitem{20}
Gu S, Zhang L, Zuo W, et al. Weighted nuclear norm minimization with application to image denoising[C]//Proceedings of the IEEE Conference on Computer Vision and Pattern Recognition. 2014: 2862-2869.
\bibitem{21}
Gu S, Xie Q, Meng D, et al. Weighted nuclear norm minimization and its applications to low level vision[J]. International Journal of Computer Vision, 2016: 1-26.
\bibitem{22}
Daubechies I, Defrise M, De Mol C. An iterative thresholding algorithm for linear inverse problems with a sparsity constraint[J]. Communications on pure and applied mathematics, 2004, 57(11): 1413-1457.
\bibitem{23}
Zhang X, Burger M, Bresson X, et al. Bregmanized nonlocal regularization for deconvolution and sparse reconstruction[J]. SIAM Journal on Imaging Sciences, 2010, 3(3): 253-276.
\bibitem{24}
Lyu Q, Lin Z, She Y, et al. A comparison of typical $\ell_p$ minimization algorithms[J]. Neurocomputing, 2013, 119: 413-424.
\bibitem{25}
Chartrand R, Wohlberg B. A nonconvex ADMM algorithm for group sparsity with sparse groups[C]//Acoustics, Speech and Signal Processing (ICASSP), 2013 IEEE International Conference on. IEEE, 2013: 6009-6013.
\bibitem{26}
Zuo W, Meng D, Zhang L, et al. A generalized iterated shrinkage algorithm for non-convex sparse coding[C]//Proceedings of the IEEE international conference on computer vision. 2013: 217-224.
\bibitem{28}
Zhang J, Zhao D, Zhao C, et al. Image compressive sensing recovery via collaborative sparsity[J]. IEEE Journal on Emerging and Selected Topics in Circuits and Systems, 2012, 2(3): 380-391.
\bibitem{29}
Zha Z, Liu X, Zhang X, et al. Compressed sensing image reconstruction via adaptive sparse nonlocal regularization[J]. The Visual Computer, 2016: 1-21.
\bibitem{30}
Larose D T. $k$-Nearest Neighbor Algorithm[J]. Discovering Knowledge in Data: An Introduction to Data Mining, 2005: 90-106.
\bibitem{31}
Wang Z, Bovik A C, Sheikh H R, et al. Image quality assessment: from error visibility to structural similarity[J]. IEEE transactions on image processing, 2004, 13(4): 600-612.
\bibitem{32}
Candes E J, Wakin M B, Boyd S P. Enhancing sparsity by reweighted $\ell_1$ minimization[J]. Journal of Fourier analysis and applications, 2008, 14(5): 877-905.
\bibitem{33}
Chang S G, Yu B, Vetterli M. Adaptive wavelet thresholding for image denoising and compression[J]. IEEE Transactions on image processing, 2000, 9(9): 1532-1546.
\bibitem{34}
Dong W, Zhang L, Shi G, et al. Nonlocally centralized sparse representation for image restoration[J]. IEEE Transactions on Image Processing, 2013, 22(4): 1620-1630.
\bibitem{35}
Osher S, Burger M, Goldfarb D, et al. An iterative regularization method for total variation-based image restoration[J]. Multiscale Modeling \& Simulation, 2005, 4(2): 460-489.
\bibitem{36}
Niknejad M, Rabbani H, Babaie-Zadeh M. Image restoration using Gaussian mixture models with spatially constrained patch clustering[J]. IEEE Transactions on Image Processing, 2015, 24(11): 3624-3636.
\bibitem{37}
Liu H, Xiong R, Zhang J, et al. Image denoising via adaptive soft-thresholding based on non-local samples[C]//Proceedings of the IEEE Conference on Computer Vision and Pattern Recognition. 2015: 484-492.
\bibitem{38}
Papyan V, Elad M. Multi-scale patch-based image restoration[J]. IEEE Transactions on image processing, 2016, 25(1): 249-261.
\bibitem{39}
Arbelaez P, Maire M, Fowlkes C, et al. Contour detection and hierarchical image segmentation[J]. IEEE transactions on pattern analysis and machine intelligence, 2011, 33(5): 898-916.
\bibitem{40}
Chen Y, Pock T. Trainable nonlinear reaction diffusion: A flexible framework for fast and effective image restoration[J]. IEEE transactions on pattern analysis and machine intelligence, 2017, 39(6): 1256-1272.
\bibitem{41}
Zha Z, Liu X, Huang X, et al. Analyzing the group sparsity based on the rank minimization methods[J]. arXiv preprint arXiv:1611.08983, 2016.
\bibitem{42}
Xu Z, Chang X, Xu F, et al. $\ell_{1/2}$ regularization: A thresholding representation theory and a fast solver[J]. IEEE Transactions on neural networks and learning systems, 2012, 23(7): 1013-1027.
\bibitem{43}
Dai T, Song C B, Zhang J P, et al. PMPA: A patch-based multiscale products algorithm for image denoising[C]//Image Processing (ICIP), 2015 IEEE International Conference on. IEEE, 2015: 4406-4410.
\end{thebibliography}
}
\end{document}